\documentclass[letterpaper, 10 pt, conference]{ieeeconf}  

\IEEEoverridecommandlockouts    

\usepackage{cite}
\usepackage{amsmath,amssymb,amsfonts}
\usepackage{graphicx}
\usepackage{textcomp}
\usepackage{booktabs}          
\usepackage{makecell}   
\usepackage{tabularx}
\usepackage[margin=1in]{geometry} 
\usepackage{float} 
\usepackage{textcomp}
\usepackage{xcolor}
\usepackage{algorithm}
\usepackage{algpseudocode}
\counterwithout{table}{section}

\title{\LARGE \bf
Topological Mapping and Navigation \\
using a Monocular Camera based on AnyLoc
}

\author{Wenzheng Zhang$^{1}$, Yoshitaka Hara$^{2}$, and Sousuke Nakamura$^{3}$
\thanks{*This work was not supported by any organization}
\thanks{$^{1}$Wenzheng Zhang is with Graduate School of Science and Engineering, Hosei University, Tokyo, Japan. 
{\tt\small wenzheng.zhang.5u@stu.hosei.ac.jp}}%
\thanks{$^{2}$Yoshitaka Hara is with Future Robotics Technology Center (fuRo), Chiba Institute of Technology, Chiba, Japan.
        }%
\thanks{$^{3}$Sousuke Nakamura is with Faculty of Science and Engineering, Hosei University, Tokyo, Japan.
        }%
}
\begin{document}

\maketitle
\thispagestyle{empty}
\pagestyle{empty}

\begin{abstract}
This paper proposes a method for topological mapping and navigation using a monocular camera. Based on AnyLoc, keyframes are converted into descriptors to construct topological relationships, enabling loop detection and map building. Unlike metric maps, topological maps simplify path planning and navigation by representing environments with key nodes instead of precise coordinates. Actions for visual navigation are determined by comparing segmented images with the image associated with target nodes.
The system relies solely on a monocular camera, ensuring fast map building and navigation using key nodes. Experiments show effective loop detection and navigation in real and simulation environments without pre-training. Compared to a ResNet-based method, this approach improves success rates by 60.2\% on average while reducing time and space costs, offering a lightweight solution for robot and human navigation in various scenarios.
\end{abstract}

\section{Introduction}

In mobile robotics and autonomous driving, map building is essential for environmental understanding and navigation. Maps enable robots to localize, plan paths, and execute tasks efficiently. Maps are generally categorized into metric maps and topological maps. 

Metric maps provide high-precision spatial information by modeling environments with specific spatial coordinates, making them suitable for accurate localization and path planning \cite{thrun2005probabilistic}. Methods such as PTAM \cite{PTAM2007}, ORB-SLAM \cite{orbslam}, LSD-SLAM \cite{LSD-SLAM2014}, and DSO \cite{engel2016directsparseodometry} have significantly advanced SLAM by leveraging visual odometry techniques for precise pose estimation and environment representation. However, these approaches require significant computational resources and struggle in dynamic or unknown environments. Additionally, building and maintaining metric maps demands extensive computation, making them less practical for large-scale or dynamic scenarios. 

Topological maps, on the other hand, represent environments by describing the connectivity between locations, reducing modeling complexity and making them ideal for large-scale, dynamic environments. Humans naturally navigate in a similar way, focusing on key locations and actions rather than precise distances \cite{spatialknowledge}. Despite these advantages, constructing and navigating with topological maps remain challenging due to the difficulty in representing nodes, arcs, and paths effectively in dynamic or unknown settings. 

To address these challenges, we propose a method for topological mapping and navigation based on the visual localization framework AnyLoc \cite{keetha2023anyloc}. AnyLoc is a universal visual place recognition method that operates effectively across diverse environments without requiring retraining or fine-tuning. In our study, we utilized AnyLoc's capability to recognize image similarity. Furthermore,  by leveraging its strong generalization capabilities, our approach enables efficient map building and navigation without relying on metric information. The system comprises two main components: (1) Mapping, which generates nodes and arcs using visual similarity from AnyLoc, and (2) Navigation, which localizes targets and plans paths based on observed images and the constructed map. This method operates solely with a monocular camera and demonstrates robust performance in both simulated and real-world environments. 

Our contributions are as follows: 
\begin{itemize}
 \item No metric information dependency: Our approach achieves purely topological mapping and navigation using only a monocular camera.
 \item Enhanced generalization and success rate: Compared to traditional ResNet-based methods, our system demonstrates significantly improved generalization, with an average success rate increase of approximately 60.2\%.  
 \item Instant deployment with a single walkthrough: Our method enables rapid deployment and navigation in unknown environments. A single monocular video recording is sufficient to build the map and navigate within it, and the system can quickly adapt to environmental changes. 
\end{itemize}
\section{Related Work}

Choset and Nagatani pioneered the concept of topological SLAM, showing how accurate localization could be achieved without relying on explicit geometric positioning\cite{Choset2001}. Building on this foundation, subsequent research integrated visual data to enhance topological mapping and navigation. For instance, Fraundorfer et al. proposed image-based topological mapping methods that extract visual features to establish environmental nodes, allowing adaptable topological maps in diverse environments\cite{Fraundorfer2007}.
\begin{figure}[t]
    \centering
    \includegraphics[width=1 \columnwidth]{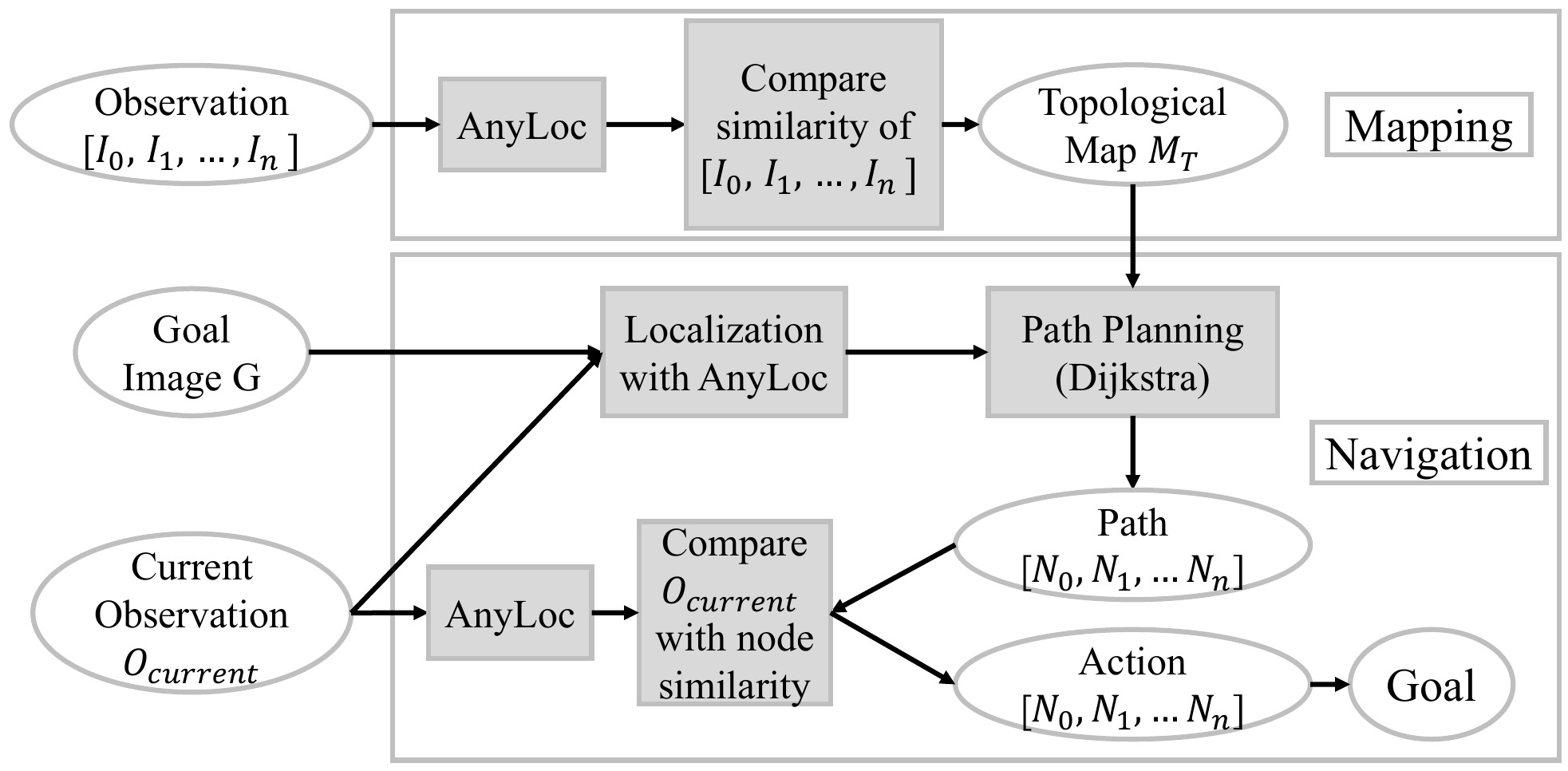} 
    \caption{Pipeline of the proposed method. From topological map building to goal navigation. There are two main components: (1) Mapping and (2) Navigation.}
    \label{fig:overview}
\end{figure}

Recent advancements in deep learning have significantly influenced topological SLAM development. Graph-based learning approaches, such as Graph Localization Networks, have improved topological localization efficiency\cite{chen2019behavioralapproachvisualnavigation}. Neural topological SLAM  further integrated neural networks with visual perception to optimize map building and navigation\cite{NeuralTopologicalSLAM2020}. Savinov et al. introduced a memory module combining parametric and non-parametric representations, enabling efficient and dynamic updates to topological structures\cite{savinov2018semiparametrictopologicalmemorynavigation}. This approach not only advanced the state-of-the-art in topological navigation but also established itself as a benchmark in the field. Meng et al. proposed a hierarchical navigation framework, combining local motion control with global topological structures, achieving efficient large-scale navigation\cite{meng2020scalinglocalcontrollargescale}. Methods like ViNG and PlaceNav have explored adaptive navigation in dynamic environments, integrating scene recognition with path planning\cite{ViNG2021,suomela2024placenavtopologicalnavigationplace}. Dhruv Shah et al. proposed General Navigation Models, advancing robot navigation toward generalization and autonomous exploration. GNM leverages large-scale multi-robot data training to achieve cross-platform zero-shot transfer\cite{shah2022gnm}. ViNT employs a Transformer-based architecture to enhance the generalization capability of vision-based navigation, demonstrating stable performance across different tasks and environments\cite{shah2023vint}. NoMaD introduces diffusion models to improve exploration efficiency and goal-directed navigation in unknown environments\cite{sridhar2024nomad}. These approaches collectively drive the evolution of robot navigation from specialized systems to general foundation models laying the groundwork for future autonomous robots.

These approaches demonstrated the potential of topological navigation but challenges remain in the field: (1) the ability to quickly construct general-purpose maps in unknown environments and (2) the dependency on extensive pre-training or environment-specific data. Our proposed approach addresses these challenges by leveraging the generalization capabilities of AnyLoc to achieve efficient, robust topological mapping and navigation in both simulated and real-world environments.

\section{Mapping and Navigation}
\subsection{Overview}
We address the problem of goal-directed visual navigation based on topological maps: given a sequence of observation images \((I_0, I_1, \dots, I_n)\), a topological map \(M_T\) is constructed. With a goal image \(G\) and the current observation \(O_{\text{current}}\), the robot determines its position and the goal position \(p_0\). Using Dijkstra's algorithm, the sequence of nodes \((N_1, N_2, \dots, N_n)\) on the topological map is selected. When navigating, a series of actions \((a_0, a_1, \dots, a_n)\) is derived by comparing the observed images \((O_1, O_2, \dots, O_n)\) with images associated with nodes \((N_1, N_2, \dots, N_n)\), enabling the robot to reach the goal \(G\), as illustrated in Fig.~\ref{fig:overview}.  

The method integrates the visual localization technique AnyLoc. By leveraging AnyLoc's image retrieval and matching capabilities, key nodes are determined, and their relationships are established to construct the topological map. AnyLoc further facilitates localization within the map and action selection, enabling navigation. 

The approach consists of two main components: (1) Mapping and (2) Navigation using the Constructed Topological Map. The following sections provide a detailed explanation of each component.

\subsection{Mapping}\label{sec:mapconstruct}
\begin{figure}[t]
    \centering
    \includegraphics[width=0.8 \columnwidth]{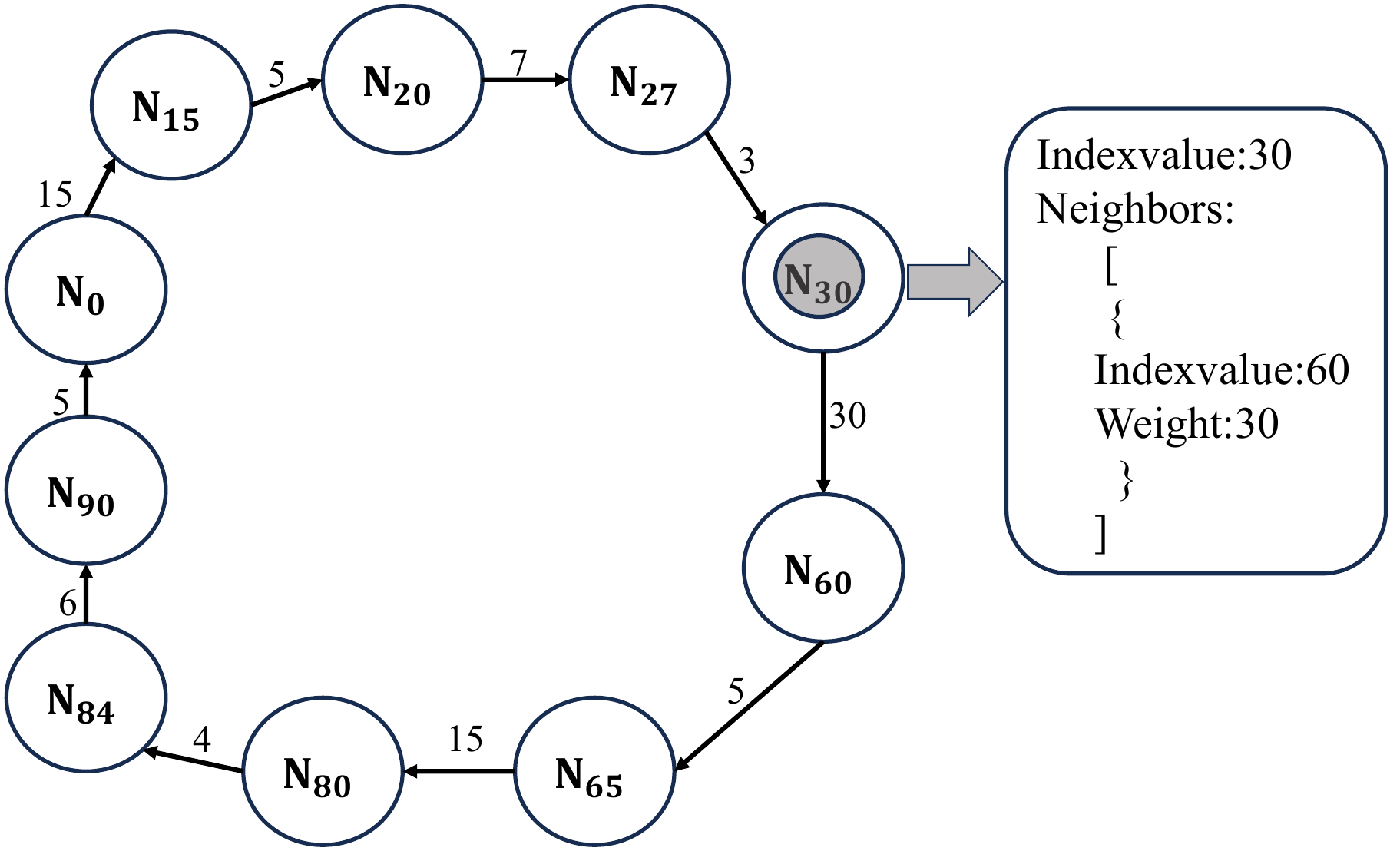} 
    \caption{Node information. The left side shows a simplified topological map, while the right side displays the information contained within each node.}
    \label{fig:Node}
\end{figure}

First, we define the concept of a topological map. A topological map is represented as an abstract directed graph composed of nodes \(N_i\) and arcs \(L_i\). Fig.~\ref{fig:Node} illustrates the information contained within the nodes and arcs. 
Each node \(N_i\) includes the following attributes: the \(indexvalue\) of the corresponding image in the database and the \(indexvalue\) of reachable nodes \(N_{i+1}\), along with the associated relative distance \(weight\). The direction of the arc indicates the connection from \(indexvalue\) to \(indexvalue\) of neighbors. 

\subsubsection{Similarity Comparison}
We utilize AnyLoc to compare the similarity between different image pairs. AnyLoc is a global descriptor-based visual feature extraction method specifically designed for visual localization tasks. It is particularly well-suited for image similarity tasks in various environments, thanks to its robust and efficient handling of diverse and complex visual data \cite{keetha2023anyloc}.
Specifically, AnyLoc employs the DINOV2 backbone to extract multi-scale visual features \cite{caron2021emergingpropertiesselfsupervisedvision}. These features are aggregated using VLAD (Vector of Locally Aggregated Descriptors) to generate compact global descriptors. The similarity score \(s\) between images is then calculated by computing the cosine similarity between their descriptors. 
The strength of AnyLoc lies in its ability to effectively handle challenges such as illumination and viewpoint variations in complex environments, providing a robust baseline for image similarity comparison tasks. 

\begin{figure}[t]
    \centering
    \includegraphics[width=1 \columnwidth]{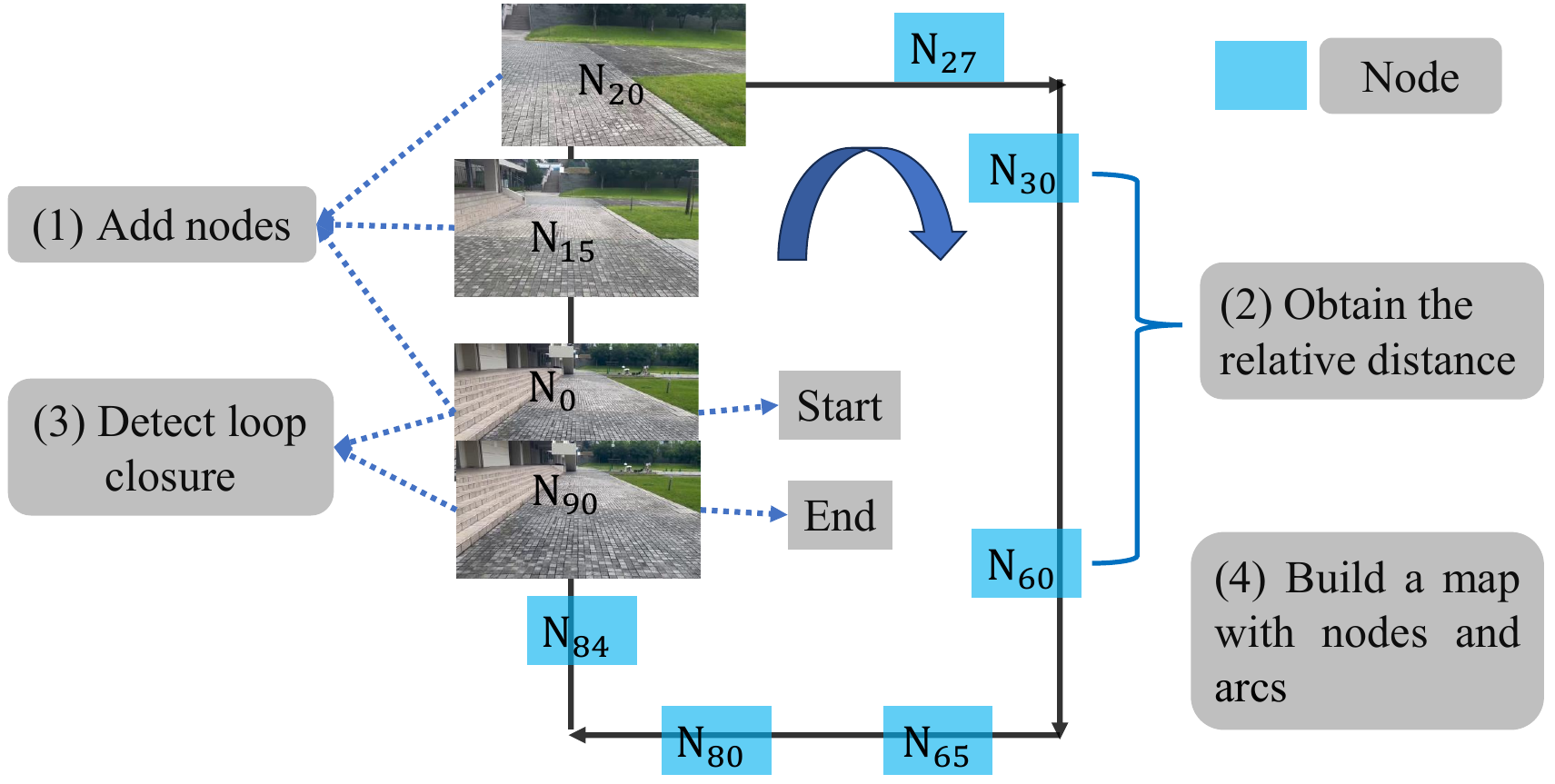} 
    \caption{Topological mapping process. Starting from \(N_0\) and moving towards \(N_{90}\) in a clockwise direction, the process of node addition, distance calculation, and loop closure detection is described.}
    \label{fig:MapConstruction}
\end{figure}
\subsubsection{Node Addition}
To construct the map, nodes must first be added, and the node addition strategy is illustrated in Fig.~\ref{fig:MapConstruction}. Before adding nodes, the collected image data \(D_i\) (where \(i\) corresponds to the frame numbers of the video segmented into images) is processed by AnyLoc to generate global descriptors. These descriptors are then stored in the corresponding nodes \(N_i\).
There are two cases when adding nodes.

Case 1: Due to the increase in distance:
Starting from \(N_0\), as the robot progresses towards \(N_{15}\), it passes through intermediate nodes from \(N_2\) to \(N_{14}\). For each intermediate node, the cosine similarity between its global descriptor and \(N_0\) is calculated, resulting in similarity scores \(s_i\). A threshold \(T_{\text{add\_new\_node}}\) is set to determine whether a new node should be added. As the distance increases, the similarity between \(N_0\) and subsequent nodes gradually decreases. When the similarity score at \(N_{15}\) drops below \(T_{\text{add\_new\_node}}\), a new node \(N_{15}\) is added to the topological map \(M_T\). 
The node addition algorithm is formally expressed in Eq.~\ref{eq:map_update}.
\begin{equation}
M_T \leftarrow 
\begin{cases} 
M_T \cup \{N_i\}, & \text{if } s_i < T_{\text{add\_new\_node}}, \\
M_T, & \text{if } s_i \geq T_{\text{add\_new\_node}}.
\end{cases}
\label{eq:map_update}
\end{equation}

Case 2: Due to significant angular changes:
From \(N_{15}\) to \(N_{20}\), the similarity score decreases gradually due to changes in orientation. As a result, \(N_{20}\) is added to the topological map \(M_T\). Similarly, nodes \(N_{27}\) to \(N_{30}\), \(N_{60}\) to \(N_{65}\), and \(N_{80}\) to \(N_{84}\) are added under similar conditions in case 2. This approach enables the automatic addition of nodes to \(M_T\) when significant changes occur.

\subsubsection{Relative Distance}
During the node addition process, a rough estimate of the relative distance, \(M_{\text{distance}}\), is calculated. When a new node \(N_i\) is added, \(M_{\text{distance}}\) is initialized to 0. A threshold, \(T_{\text{add\_distance}}\), is defined such that the distance counter only increments when the similarity score between the current frame \(D_i\) and the previous frame \(D_{i-1}\) falls below this threshold. This effectively prevents distance increments when the robot is stationary, even as the index of frames increases. The distance algorithm is formally expressed in Eq.~\ref{eq:distance_update}.
\begin{equation}
M_{\text{distance}} =
\begin{cases} 
0, & \text{added newnode}(D_i), \\
M_{\text{distance}} + 1, & \text{similarity}(D_i, D_{\text{previous}})\\ & \quad < T_{\text{add\_distance}}, \\
M_{\text{distance}}, & \text{otherwise}.
\end{cases}
\label{eq:distance_update}
\end{equation}

\subsubsection{Loop Closure Detection}
During system operation, each node \(N_i\) is compared for similarity against all nodes in the topological map \(M_T\). \(T_{\text{loop\_closure}}\) is defined to determine whether a loop closure has occurred. If the similarity score \(s_i\) exceeds \(T_{\text{loop\_closure}}\), a loop closure is detected.

As illustrated in Fig.~\ref{fig:MapConstruction}, when the robot reaches \(N_{90}\) from \(N_{84}\), a new node \(N_{90}\) would typically be added due to the conditions in Case 1. However, since  \(s_{90}\) between \(N_{90}\) and an existing node \(N_{0}\) exceeds \(T_{\text{loop\_closure}}\), \(N_{90}\) is not added to \(M_T\). Instead, \(N_{84}\) is connected directly to \(N_{0}\), and the relative distance \(M_{\text{distance}}\) is updated accordingly.
This process effectively prevents redundant or nodes from being added to \(M_T\), ensuring a more compact and efficient topological map. 

\subsubsection{Map Optimization}
\(T_{\text{interval}}\) is defined to determine the shortest interval between the adding node and the previous node. By adjusting \(T_{\text{interval}}\), \(T_{\text{add\_node}}\), and \(T_{\text{loop\_closure}}\), the sparsity of the map can be controlled, and unnecessary nodes can be removed to optimize the map. 

\begin{figure}[t]
    \centering
    \includegraphics[width=0.9 \columnwidth]{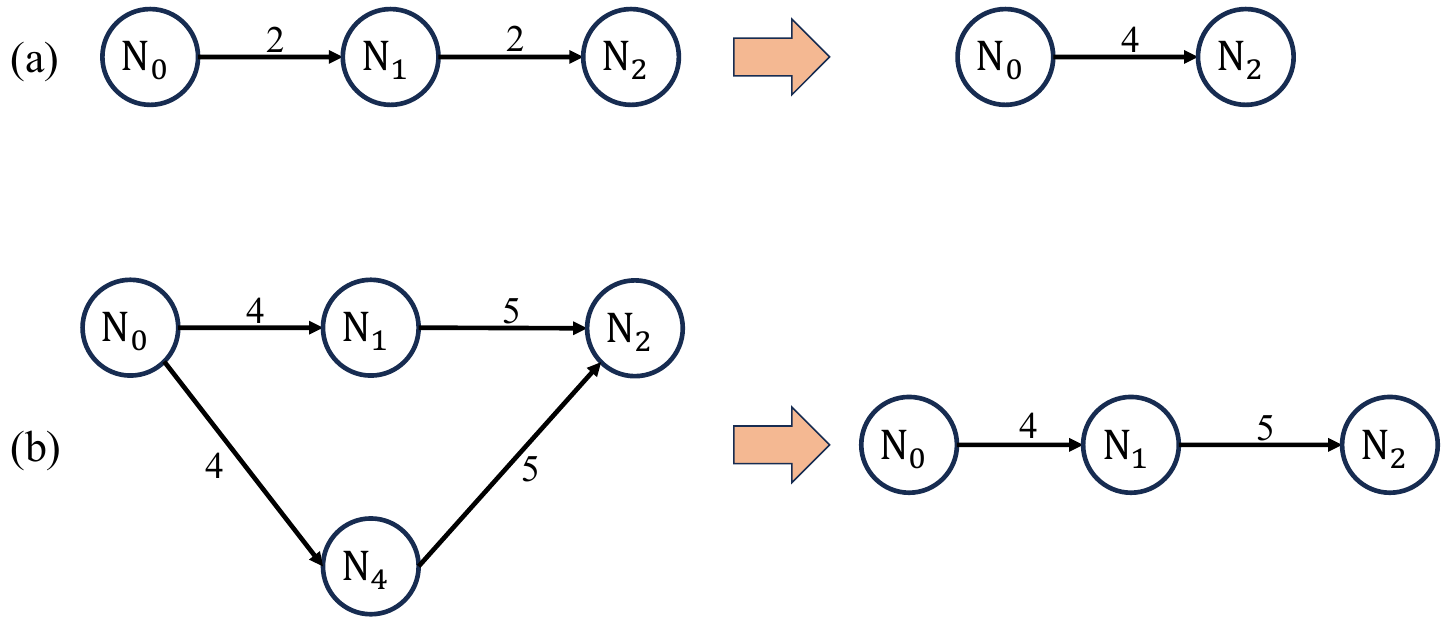} 
    \caption{Topological map node optimization. In case (a), \(N_1\) is close to both \(N_0\) and \(N_2\). In case (b), nodes \(N_1\) and \(N_4\) are similar. The right side of the image shows the optimized results.}
    \label{fig:nodeoptimiztion}
\end{figure}

As shown in Fig.~\ref{fig:nodeoptimiztion}, in case (a), increasing \(T_{\text{interval}}\) or decreasing \(T_{\text{add\_node}}\) makes the topological map \(M_T\) more sparse. 
In case (b), nodes \(N_1\) and \(N_4\) are similar, and \(N_4\) corresponds to a location already represented by \(N_1\), so \(N_4\) should not be added to \(M_T\). By decreasing \(T_{\text{loop\_closure}}\), \(N_4\) can be identified as a loop closure and merged with \(N_1\). The arcs are then updated, resulting in the optimized map shown on the right side of Fig.~\ref{fig:nodeoptimiztion}.

\subsection{Navigation}
\begin{figure}[t]
    \centering
    \includegraphics[width=1 \columnwidth]{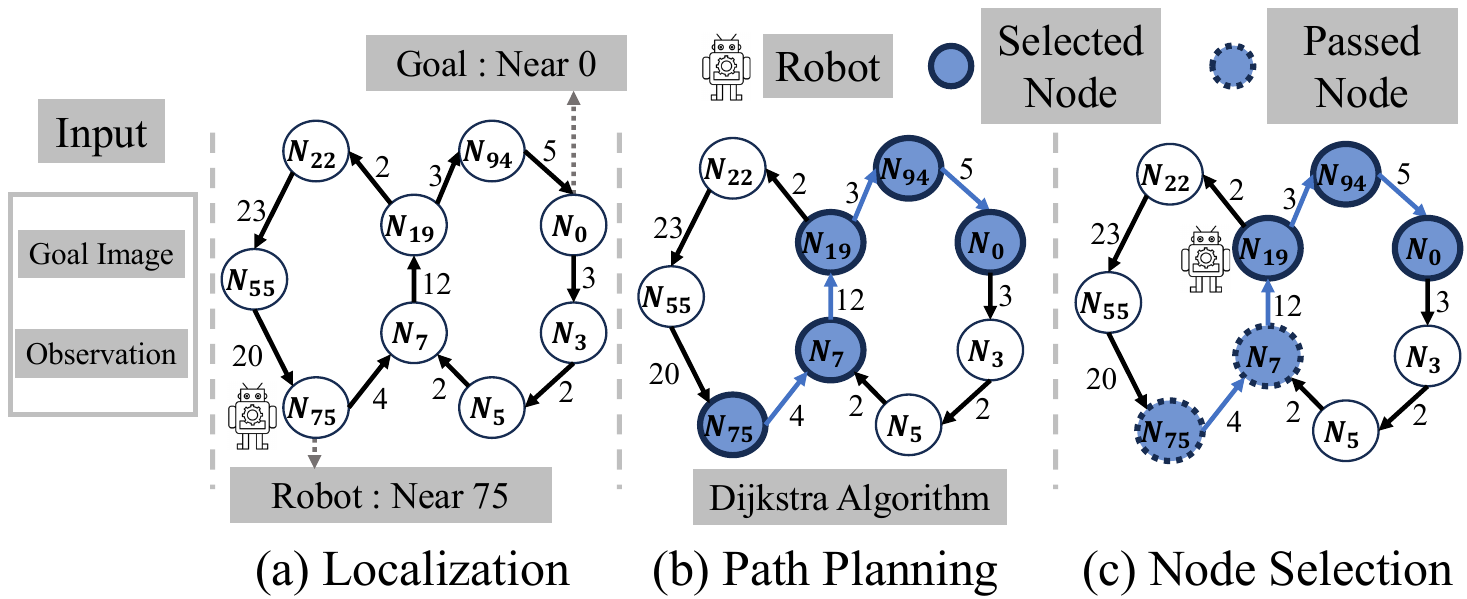} 
    \caption{Global navigation system overview. (a) Locate nodes \(N_{75}\) and \(N_0\) in the graph, corresponding to the current observation and the target image, respectively. (b) The shortest path on the graph between these nodes is selected. (c) Node selection is performed during the movement towards the goal.}
    \label{fig:navigation}
\end{figure}
\(Navigation\) consists of four components. Fig.~\ref{fig:navigation} illustrates three components in the global level, while the other one is \(Action\ Control\) in the local level. The inputs are the goal image \(G\) and the robot's current observation \(O_{\text{current}}\). The process begins with \(Localization\), which determines the positions of the goal and the robot. Next, \(Path\ Planning\) uses Dijkstra's algorithm to generate a navigation sequence \(L_0, L_1, \dots, L_n\). The \(Node\ Selection\) and \(Action\ Control\) components ensure local control and sequential navigation through the nodes, with \(AnyLoc\) assisting in \(Localization\). This process enables the robot to reach the \(G\).
\subsubsection{Localization}
Localization is crucial for the entire system, as it underpins three key components of the navigation process. A threshold \(T_{\text{milestone}}\) is set to ensure localization performance.
There are two types localization in system.
Initial Localization:  
   Using AnyLoc, global descriptors are generated for the input \(G\) and \(O_{\text{current}}\). These descriptors are compared against those stored in \(M_T\) for each node \(N_i\) using cosine similarity to produce a sequence of similarity scores \(s_i\). If \(s_i\) exceeds \(T_{\text{milestone}}\), the node \(N_i\) with the highest \(s_i\) is selected as the localization node. If no such \(s_i\) is found, the robot rotates counterclockwise and attempts localization again until a similarity score exceeding \(T_{\text{milestone}}\) is achieved. As shown in Fig.~\ref{fig:navigation} (a), the robot localizes near \(N_{75}\), while the target is located near \(N_0\).  

General Localization:  
   The robot continuously updates its position by generating global descriptors for the current observation \(O_{\text{current}}\) and repeating this process to dynamically track its position during navigation.

\subsubsection{Path Planning}
Using the two nodes obtained during initialization as inputs, the navigation sequence \(L_0, L_1, \dots, L_n\) is generated based on the relationships between nodes and arcs in \(M_T\) using Dijkstra's algorithm. This sequence corresponds to selected nodes \(N_{i}\), as shown in Fig.~\ref{fig:navigation} (b). Subsequent navigation proceeds by sequentially switching through the nodes in the navigation sequence.
\subsubsection{Node Selection}
Using the previously described localization method, the robot can identify the node closest to the current navigation sequence. The robot's position can only be confirmed near a specific node if the similarity score \(s_i\) exceeds \(T_{\text{milestone}}\). Additionally, a node-switching threshold \(T_{\text{change\_node}}\) is set. As the robot moves from the current node \(L_i\) toward the next node \(L_{i+1}\), the similarity score \(s_i\) between \(L_i\) and \(O_{\text{current}}\) gradually decreases. When \(s_i\) drops below \(T_{\text{change\_node}}\), the robot switches to \(L_{i+1}\), with a new similarity score \(s_{i+1}\). 
If \(s_{i+1}\) exceeds \(T_{\text{milestone}}\) and \(i\) is less than the length of navigation sequence minus one, then the robot is considered to have successfully reached \(L_{i+1}\). When \(i\) equals the length of navigation sequence minus one, the robot is deemed to have reached the goal node. 
To avoid mismatches caused by matching nodes that are too far ahead or behind in the navigation sequence, boundaries are set during the \(Localization\) process to limit the range of nodes that can be matched. These boundaries ensure that the robot can only match nodes within the vicinity of the current node.
\subsubsection{Local Motion Control}
\begin{figure}[t]
    \centering
    \includegraphics[width=0.9 \columnwidth]{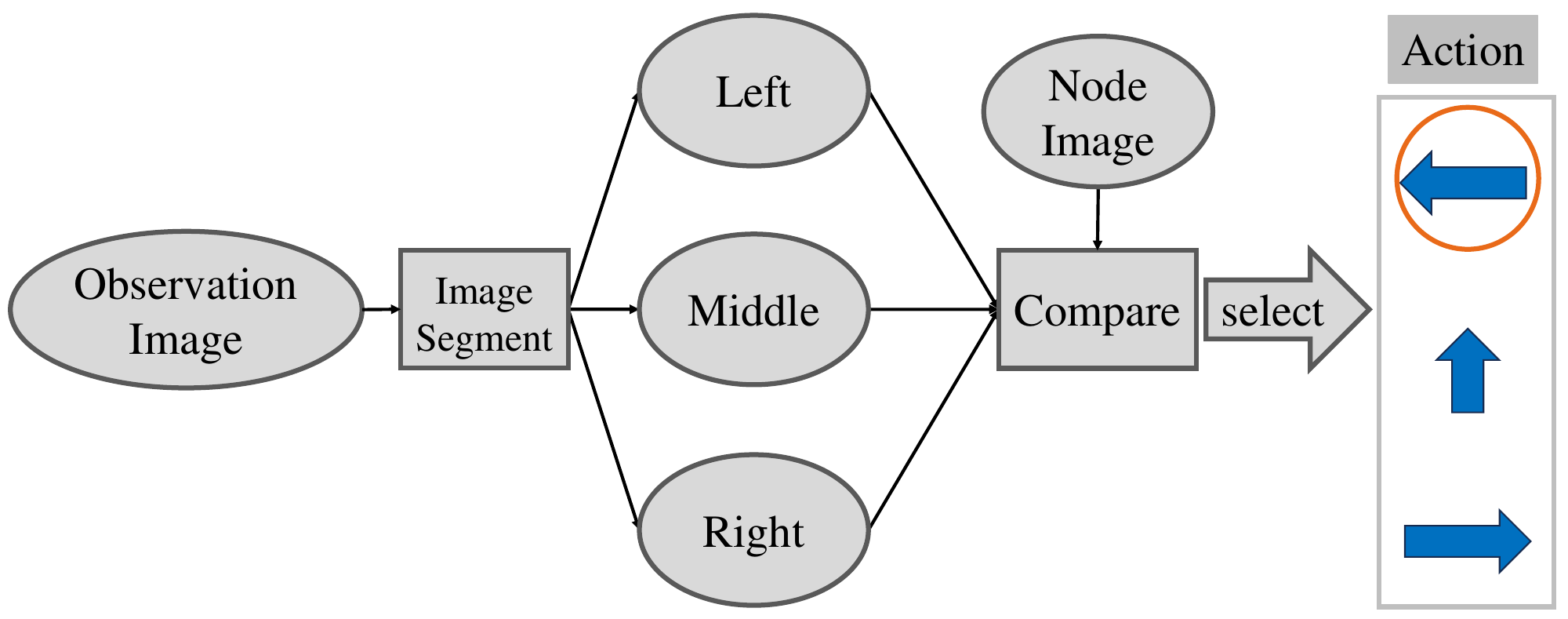} 
    \caption{Action control. Given an observation image, it is divided into three parts to generate three descriptors, each of which is compared with the node's descriptor. The most suitable action is then selected based on the comparison.}
    \label{fig:action control}
\end{figure}

To enable movement between nodes, the robot performs motion control based on the current observation \(O_{\text{current}}\), as shown in Fig.~\ref{fig:action control}. The image is segmented into three parts—\(Left\), \(Middle\), and \(Right\)—and compared with the \(Node\ Image\) using AnyLoc to generate similarity scores. If the highest similarity \(s_{\text{highest}}\) exceeds the threshold \(T_{\text{limited\_control}}\), the robot moves forward, left, or right, corresponding to the \(Middle\), \(Left\), or \(Right\) segments, respectively. A two-cycle process ensures accuracy: in the first cycle, the action is stored; in the second, it is executed if both cycles select the same action. If a conflict occurs, no action is taken unless \(Middle\) is selected in a certain cycle. In this case, the motion combines forward movement with a left or right turn, resulting in forward-left or forward-right movement as needed. This approach ensures smoother and more reliable control.
\section{Experiments}
\subsection{Overview}
We conducted experiments in both real and simulation environments and compared our approach with previous work. This study addresses the following questions:  \\
\(\textbf{Q1}\): Can mapping and navigation be effectively achieved in different environments without pre-training?  \\
\(\textbf{Q2}\): How does the proposed method perform compared to previous baseline methods?\\
\(\textbf{Q3}\): What is the impact of topological map sparsity on topological navigation?
\subsection{Experiment Environments}
\subsubsection{Real-World Environment}
\begin{figure}[t]
    \centering
    \includegraphics[width=1 \columnwidth]{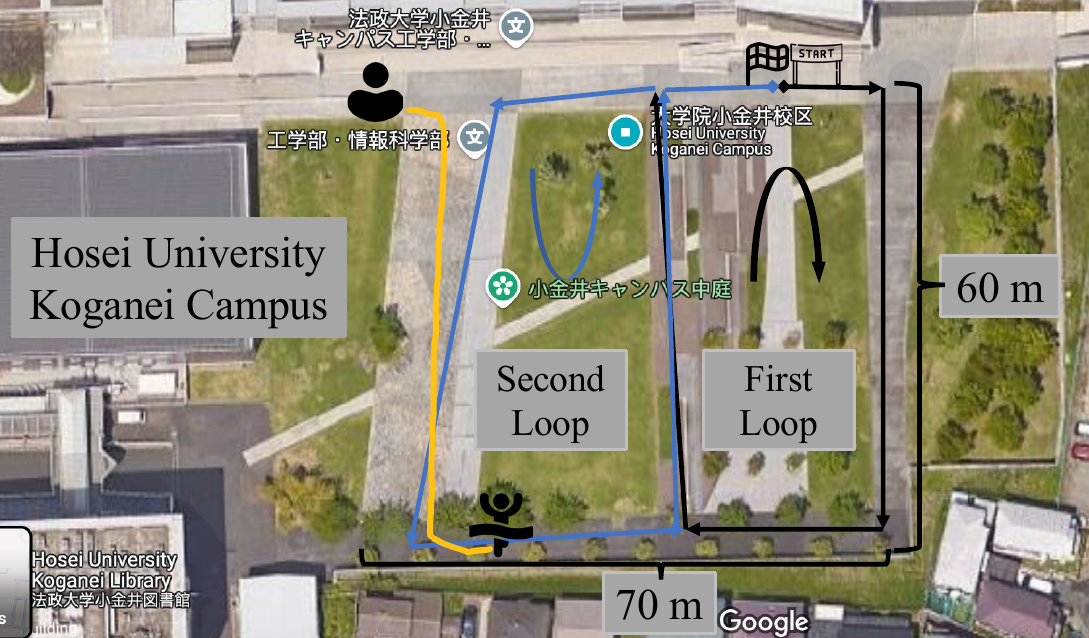} 
    \caption{Real environment. The black and blue lines represent the data collection routes, while the orange line represents the outdoor navigation test route.}
    \label{fig:real environment}
\end{figure}

The experiments were conducted in the courtyard of Hosei University's Koganei Campus, covering an area approximately 70 meters in length and 60 meters in width, as illustrated in Fig.~\ref{fig:real environment}. The experimental environment features sidewalks, green spaces, and diverse vegetation, occasionally traversed by pedestrians.
\subsubsection{Simulation Environment}
\begin{figure}[t]
    \centering
    \includegraphics[width=1 \columnwidth]{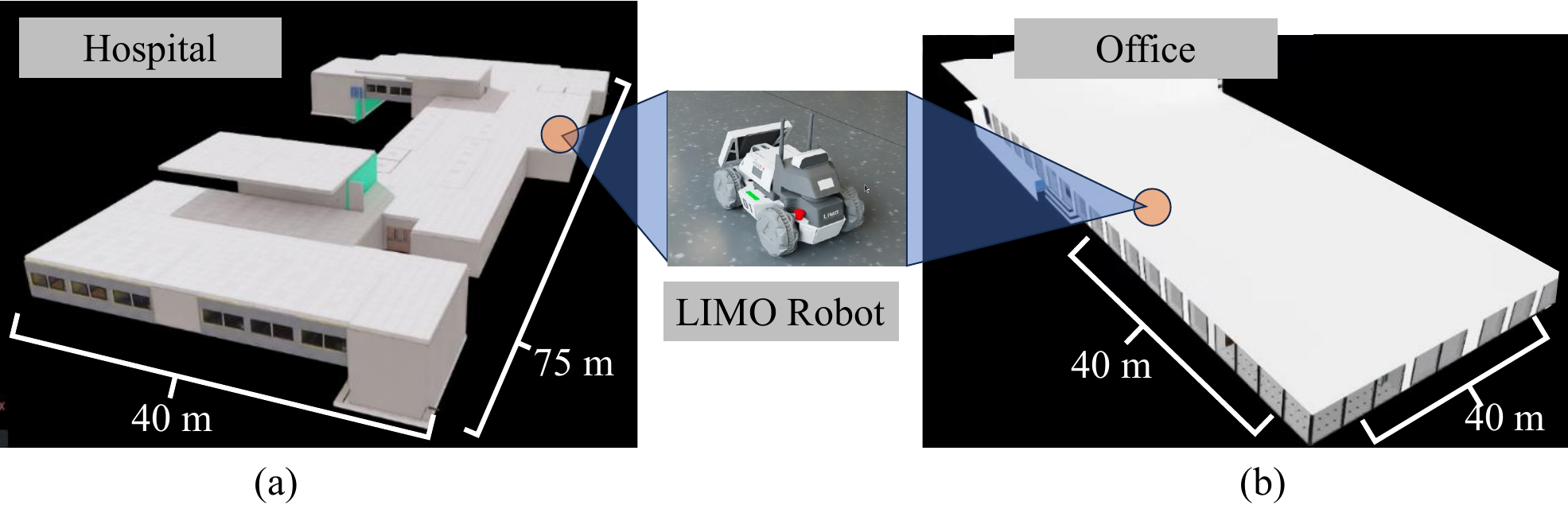} 
    \caption{Simulation environment. (a) and (b) depict the external views and layouts of the hospital and office environments, respectively, with the approximate robot positions shown.}
    \label{fig:Simulation environment}
\end{figure}

This research utilized the NVIDIA Isaac Sim platform for simulation experiments \cite{isaacsim}. Isaac Sim is a high-fidelity robotics simulation tool based on NVIDIA Omniverse, supporting physics simulation, camera sensor modeling, and integration with ROS/ROS2. We utilized two simulated environments provided by Isaac Sim. The experimental environment in Fig.\ref{fig:Simulation environment}(a) represents a hospital setting. In addition, to evaluate the generalizability of our approach, we conducted experiments in a simulated office environment, as shown in Fig.\ref{fig:Simulation environment}(b). This environment is comparatively brighter than the hospital setting. In the simulation, the Omniverse Kit interface of Isaac Sim was integrated with ROS2, enabling real-time control and data collection. The robot model used was Limo, equipped with a monocular camera with a horizontal field of view (HFOV) of 65° for data acquisition and visual navigation.

\subsection{Experiment Methodology}
All data processing and algorithm execution were conducted on a desktop computer equipped with an NVIDIA GeForce GTX 4070 Ti SUPER GPU and an Intel i7-13700 CPU. 

\subsubsection{Real-World Environment}
Data Collection: Data was collected in the experimental area using an iPhone 13 Pro Max, recording at 1080p with 30 frames per second. As shown in Fig.~\ref{fig:real environment}, the data collection started at the \(start\) position, proceeding clockwise along the black line, and then counterclockwise along the blue line. The overlap between the black and blue lines indicates the occurrence of loop closure. The route concluded near the goal flag. The collected video was processed on the desktop computer, saving frames at intervals of 30.  

\begin{figure}[t]
    \centering
    \includegraphics[width=0.9 \columnwidth]{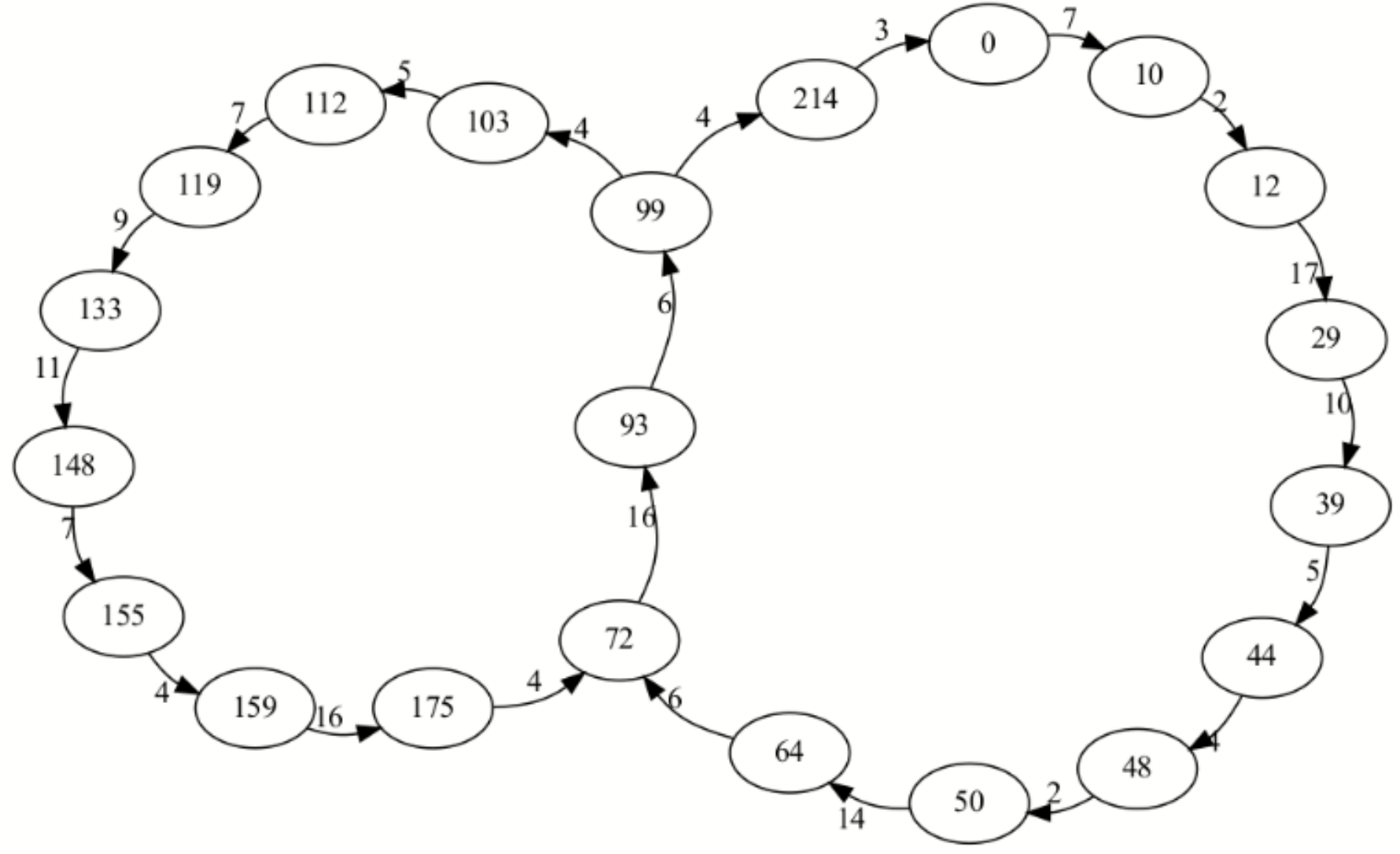} 
    \caption{Topological map of the real environment. }
    \label{fig:topologicalmap}
\end{figure}

Map building: A topological map was constructed using the method described in Section \ref{sec:mapconstruct}. The sparsity of the map was controlled through parameter adjustments, and the resulting map is shown in Fig.~\ref{fig:topologicalmap}.  

Navigation Experiment: Experiments were conducted at different times to evaluate the generalizability of the method. Starting from the upper-left corner in Fig.~\ref{fig:real environment}, marked by the small human icon, a person followed the yellow trajectory to stop at the lower human icon. A target image was provided as input, and the collected video was replayed to test whether the algorithm could accurately identify the current node, match the target image to its corresponding node, generate a correct navigation sequence, and guide the robot to move in the proper direction. The experiment observed whether the robot could accurately reach and recognize the target.  

\subsubsection{Simulation Environment}
Data Collection: Data was collected in the experimental area by navigating the robot along predefined paths. Images were captured using ROS2 during the robot's movement.  

Map building: The map creation process was identical to that in the real-world environment.  

Navigation Experiment: Target images or node indices were provided as input, and the algorithm was tested for its ability to navigate the robot along the planned route on the topological map and successfully achieve topological navigation.
\begin{figure}[t]
    \centering
    \includegraphics[width=1 \columnwidth]{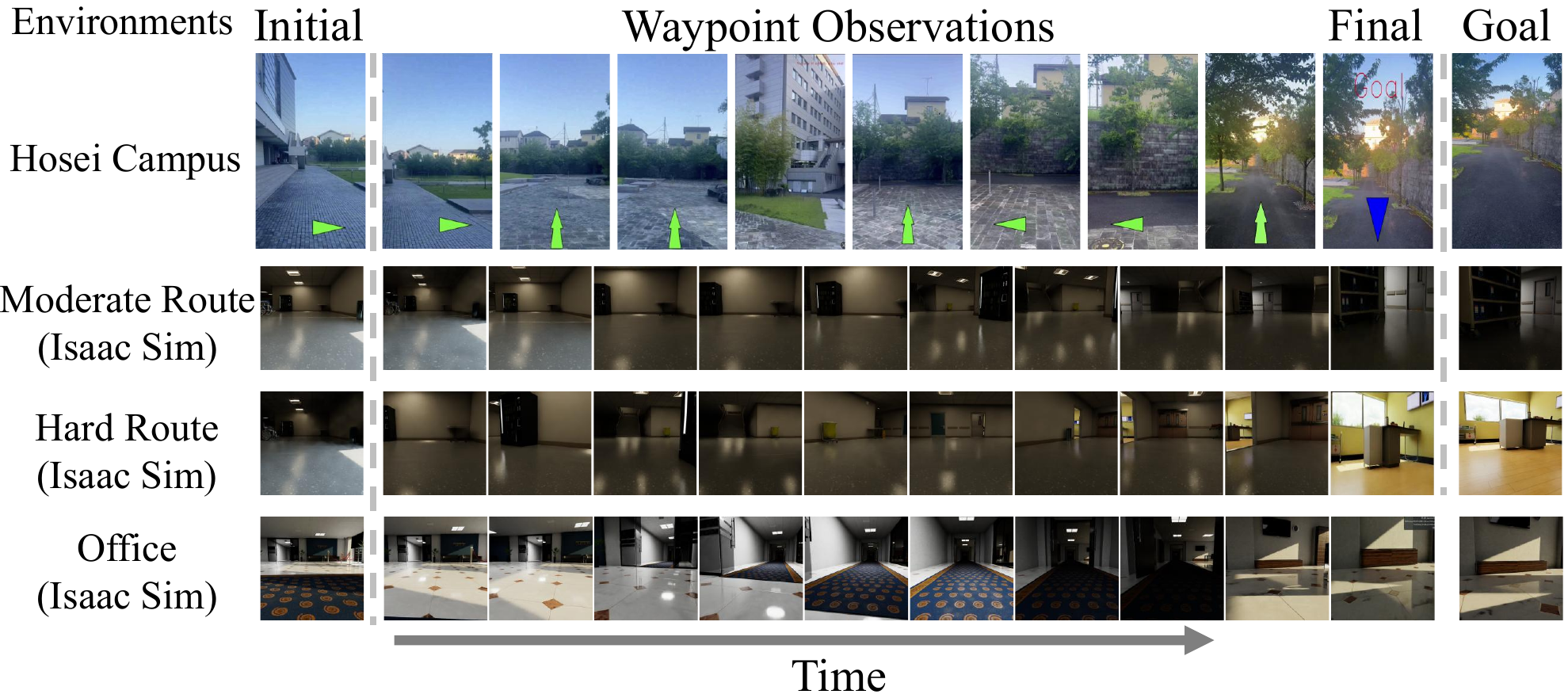} 
    \caption{Generalization experiments: We evaluate our model in different environments. The first column represents the experimental environments. Given a goal image (last column), the robot or a person attempts to navigate to the goal. Intermediate columns show observation images from visited nodes. The penultimate column shows the final observation, demonstrating that the robot successfully reached the goal, it can be compared with the goal images in the last column.}
    \label{fig:Generalization Experiments}
\end{figure}
\subsection{Evaluation Methodology}  
We evaluated our method in simulation following \cite{anderson2018evaluationembodiednavigationagents}, comparing its \(success \ rate\ (SR)\) with the baseline SPTM \cite{savinov2018semiparametrictopologicalmemorynavigation}. The SR is the ratio of successful attempts to total attempts on a test route, indicating whether the robot reaches the target. This indicator is critical for assessing navigation performance. 

We evaluated routes of different lengths in the hospital environment—10 m, 20 m, and 35 m—and classified routes into three difficulty levels: \(easy\), \(moderate\), and \(hard\), based on the number of turns and obstacles. Moreover, a route of approximately 40 meters in length was established in the office environment and classified as moderate in terms of difficulty. This evaluation framework allowed a comprehensive comparison with the baseline method, highlighting the strengths and limitations of the proposed approach.  
Following the SPTM methodology, data was collected through exploration in the environment, and a ResNet18 model was used via transfer learning to train a custom action prediction model, enhancing its competitiveness. Moreover, in the office environment, the ResNet-based method used the model trained in the hospital environment to evaluate generalizability. To ensure the accuracy of the comparison, both methods utilized AnyLoc for localization. Each model was tested 14 times on each route for a fair and consistent evaluation.
\subsection{Results}
As shown in Fig.~\ref{fig:Generalization Experiments}, our method successfully achieves navigation in both real and virtual environments, indoors and outdoors, with a minimal number of nodes and without the need for pre-training. This addresses \(Q1\).  

Furthermore, as indicated in Table~\ref{tab:comparison}, our method consistently outperforms the baseline across all test routes, and this advantage becomes increasingly evident as the difficulty and length of the routes increase. Meanwhile, evaluations conducted in the office environment further validate the generalizability of our model, achieving a 75\% improvement in performance and a 60.2\% increase in overall average success rate, thereby providing a clear response to \(Q2\).

\begin{table}[t]
\caption{Comparison of Success Rates\\ by Methods and Difficulty Levels}
\label{tab:comparison}
\begin{tabular}
{@{}>{\centering\arraybackslash}m{1.6cm}>
{\centering\arraybackslash}m{0.9cm}>
{\centering\arraybackslash}m{0.9cm}>
{\centering\arraybackslash}m{0.8cm}>
{\centering\arraybackslash}m{0.65cm}>
{\centering\arraybackslash}m{1.0cm}@{}} 
\toprule
\textbf{Method} & \textbf{Easy 10 m} & \textbf{Moderate 20 m} & \textbf{Hard 35 m} & \textbf{Office 40 m} & \textbf{Average} \\
\midrule
\begin{tabular}[c]{@{}c@{}}ResNet+AnyLoc\end{tabular} & 1.00 & 0.64 & 0.47 & 0.25 & 0.59  \\
Ours & 1.00 & 0.92 & 0.86 & 1.00 & 0.95 \\

\bottomrule
\end{tabular}
\end{table}
\begin{figure}[t]
    \centering
    \includegraphics[width=1 \columnwidth]{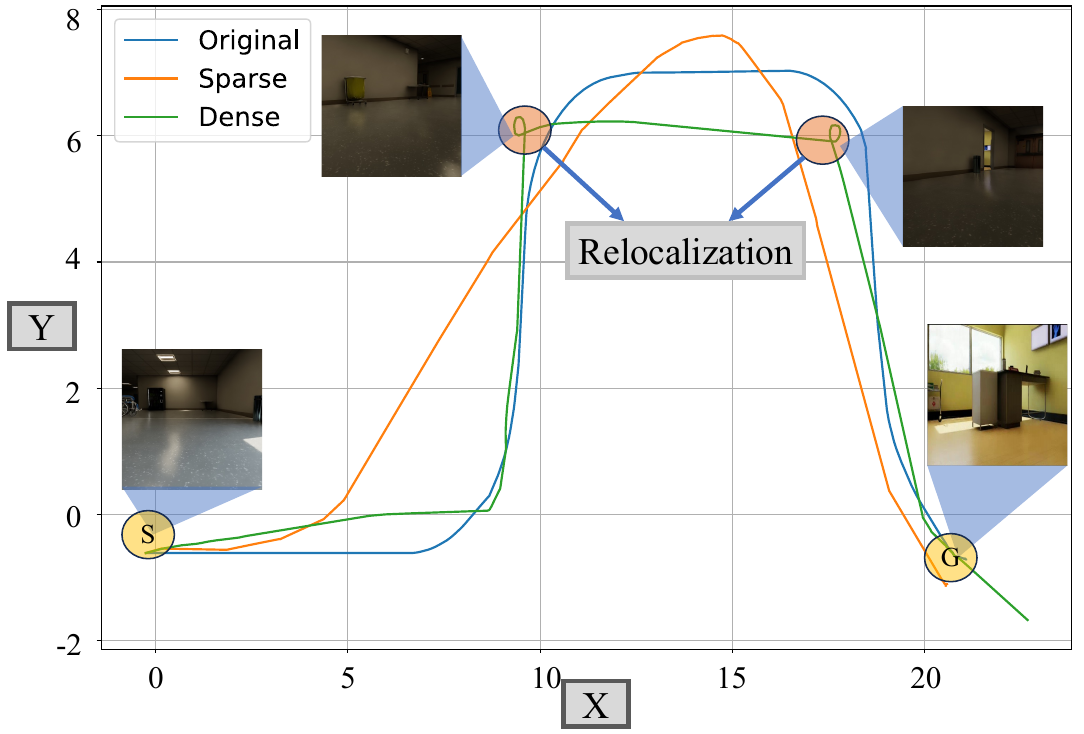}
    \caption{Impact of topological map sparsity on navigation. S is the start point, and G is the goal point. The blue line represents the robot's original trajectory during data collection, while the orange and green lines represent the trajectories on the topological map when it is sparse and dense, respectively.}
    \label{fig:Sparsity on Navigation}
\end{figure}
To address \(Q3\), we conducted navigation experiments using maps with varying levels of sparsity, as shown in Fig.~\ref{fig:Sparsity on Navigation}. When the nodes in the map are placed too closely together, the navigation trajectory adheres more closely to the original path but is more prone to re-localization events. In contrast, sparser maps significantly reduce the occurrence of re-localization. This is because overly dense maps increase the likelihood of the robot missing navigation nodes during forward motion, leading to frequent re-localization attempts. Therefore, denser nodes in the topological map do not necessarily lead to better results. An appropriate level of sparsity is crucial for effective topological navigation.
\section{Conclusion}

In this paper, we introduced a system for topological mapping and goal-directed navigation using a monocular camera in both real and simulated environments. By incorporating AnyLoc, the system efficiently selects key nodes based on the variation in image similarity, enabling rapid construction of topological maps. During visual navigation, AnyLoc was employed for node localization and action selection. Through extensive experiments, we addressed three key research questions and demonstrated that our method outperforms baseline approaches across various routes. Notably, our system achieves fast map building and visual navigation in new environments without requiring pre-training.

Current limitations include the need for manual parameter tuning during map building and navigation. Progress has been made in addressing the issue of parameter adjustment. We plan to complete the integration work in the future. Additionally, we aim to incorporate richer information from observation images to further  enhance the robustness and reliability of the navigation process.
\addtolength{\textheight}{-4cm}   


\bibliographystyle{IEEEtran}
\bibliography{refer}

\end{document}